\documentclass{article}

\usepackage{arxiv}

\usepackage[utf8]{inputenc} 
\usepackage[T1]{fontenc}    
\usepackage{hyperref}       
\usepackage{url}            
\usepackage{booktabs}       
\usepackage{nicefrac}       
\usepackage{microtype}      
\usepackage{lipsum}		
\usepackage{graphicx}
\usepackage{natbib}
\usepackage{doi}
\usepackage{bbm}
\usepackage{amsmath}

\title{Cascade Learning Localises Discriminant Features in Visual Scene Classification}

\author{Junwen Wang\thanks{junwenwang86@gmail.com} \\
	School of Electronics \& Computer Science\\
	University of Southampton\\
	Southampton, UK \\
        SO17 3AT \\
	\And
	Katayoun Farrahi \\
	School of Electronics \& Computer Science\\
	University of Southampton\\
	Southampton, UK \\
        SO17 3AT \\
}

\date{}

\renewcommand{\shorttitle}{\textit{arXiv} Template}

\hypersetup{
pdftitle={A template for the arxiv style},
pdfsubject={q-bio.NC, q-bio.QM},
pdfauthor={David S.~Hippocampus, Elias D.~Striatum},
pdfkeywords={First keyword, Second keyword, More},
}

\begin{document}
\maketitle

\begin{abstract}
Lack of interpretability of deep convolutional neural networks (DCNN) is a well-known problem particularly in the medical domain as clinicians want trustworthy automated decisions. One way to improve trust is to demonstrate the localisation of feature representations with respect to expert-labeled regions of interest. In this work, we investigate the localisation of features learned via two varied learning paradigms and demonstrate the superiority of one learning approach with respect to localisation. Our analysis on medical and natural datasets shows that the traditional end-to-end (E2E) learning strategy has a limited ability to localise discriminative features across multiple network layers. We show that a layer-wise learning strategy, namely cascade learning (CL), results in more localised features. Considering localisation accuracy, we not only show that CL outperforms E2E but that it is a promising method of predicting regions. On the YOLO object detection framework, our best result shows that CL outperforms the E2E scheme by $2\%$ in mAP.
\end{abstract}

\section{Introduction}
\label{sec:introduction}
Deep Learning (DL) advances in computer vision~\cite{10.5555/2999134.2999257,9356353, Redmon2018YOLOv3AI} have been successfully applied to specialist domains such as medical imaging~\cite{Esteva2021}, improving performance in pathology detection from chest radiograph~\cite{Irvin2019, Arias-Londono2020}, finding malignant lesions from skin scans~\cite{Liu2020} and predicting patient survival from whole slide images~\cite{SRINIDHI2021101813}. The success of DL in medical imaging motivates further investigation into feature understanding as these architectures suffer from their black-box nature, raising valid concerns by medical practitioners. 

Interpretability is generally the ability for a human to understand the reasons (i.e. features) behind the decision made by the system. Simple machine learning models, such as logistic regression or decision trees, are more easily interpretable though do not perform nearly as well as DCNNs with millions of parameters. Feature visualisation~\cite{reyes2020interpretability} is the current state of the art approach for DCNN interpretation. 
Feature visualisation techniques generate localisation maps, highlighting the pixels and regions in the input image used in making the prediction~\cite{Saporta2021.02.28.21252634, simonyan2014deep, Selvaraju2020}.

Cascade Learning ({\bf CL})~\cite{marquez2018deep}, which builds on the idea of the cascade correlation algorithm~\cite{10.5555/109230.107380}, is an alternative way of training a DCNN. This learning paradigm differs from traditional end-to-end (E2E) learning, whereby all of the layers of the network are learned simultaneously, resulting in varied feature representations. Recent studies~\cite{du2019transfer, mlhc2022} demonstrate the superior performance of transferring CL features to downstream classification tasks. In this paper, we investigate the difference in feature representations considering localisation as a key metric for traditional E2E learning versus CL. We observe that CL does result in more localised features, considering several metrics and visualisation approaches, and these features appear to be more localised at every layer of the DCNN.

We then take these findings one step further and consider whether the improved feature localisation results in superior object detection. Object detection frameworks train an effective bounding box regressor to classify and localise the object in an image or video~\cite{Redmon2016YouOL, Redmon2018YOLOv3AI}. In this work, we consider the association between visually localised features and the bounding box prediction. We seek to answer: {\em does the superior localisation ability of CL further improve the ability of the model to predict the bounding box region of interest?} We find that CL is promising and improves bounding box region of interest predictions in comparison to the widely adopted E2E training scheme. 

\begin{figure}[!t]
    \centering
    \includegraphics[width=0.7\textwidth]{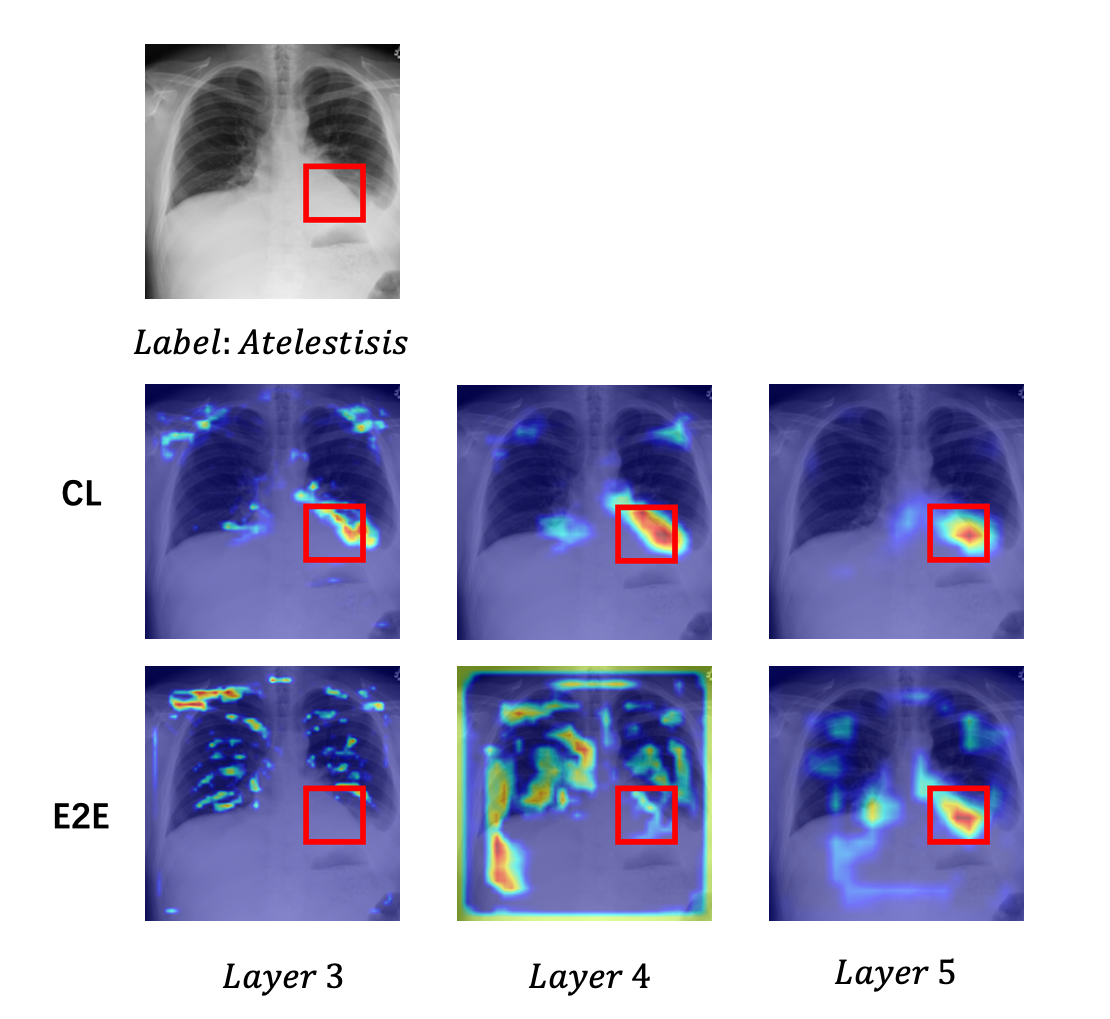}
    \caption[Grad-CAM Saliency Map visualisation at different layers]{Grad-CAM saliency map visualisation at different layers of the neural network. Results on a (top) cascade-trained network versus (bottom) E2E training. By comparing the features to the red rectangle denoting the bounding box, CL achieves better localisation.}
    \label{fig:gradcam_chestxray}
\end{figure}

\begin{figure}[!t]
    \centering
    \begin{tabular}{c}
    Original \\
    \includegraphics[width=3cm,height=3cm]{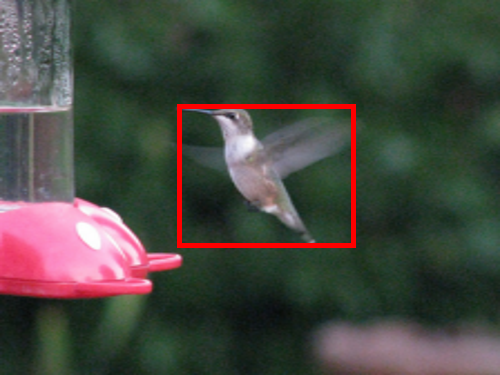}\\
    \small (a) \textit{Bird}
    \end{tabular}
    \begin{tabular}{c}
    CL \\
    \includegraphics[width=3cm,height=3cm]{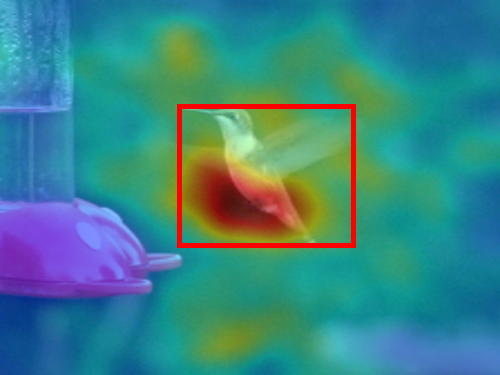}\\
    \small (b) \textit{$IOU = 0.22$}
    \end{tabular}
    \begin{tabular}{c}
    E2E \\
    \includegraphics[width=3cm,height=3cm]{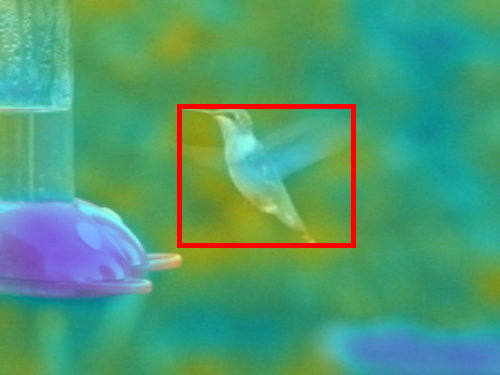}\\
    \small (c) \textit{$IOU = 0.12$}
    \end{tabular}
    
    \begin{tabular}{c}
    \includegraphics[width=3cm,height=3cm]{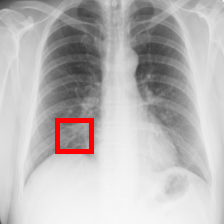}\\
    \small (d) \textit{Atelectasis}
    \end{tabular}
    \begin{tabular}{c}
    \includegraphics[width=3cm,height=3cm]{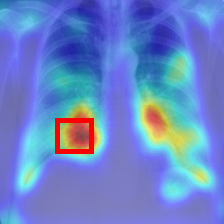}\\
    \small (e) \textit{$IOU = 0.17$}
    \end{tabular}
    \begin{tabular}{c}
    \includegraphics[width=3cm,height=3cm]{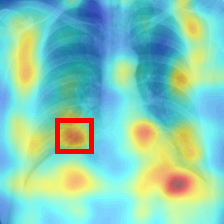}\\
    \small (f) \textit{$IOU = 0.05$}
    \end{tabular}
    
    \caption{Saliency map generated via CL in comparison to the same network which is E2E trained. Left column: Original image and its corresponding label; Middle: \textbf{CL}; Right: \textbf{E2E}. The heatmap was generated after post-processing using a Gaussian filter.}
    \label{fig:CL_saliency}
\end{figure}

The main contributions of this paper are as follows: 
\begin{itemize}
     \item Our analysis via various feature visualisation techniques shows that traditional E2E training has a limited ability to localise discriminative features across the intermediary layers of a DCNN.
     \item We demonstrate that using a layer-wise learning strategy, namely cascade learning, leads to an improvement in feature localisation.
     \item Quantifying the degree of overlap between the binarized mask and the bounding box, for the Chest X-ray dataset, $86\%$ images have more localised features, with CL showing a consistent improvement across every network layer.
     \item We find the superior localisation ability leads to further improvement in predicting bounding box regions of interest. Our bounding box prediction via CL trained backbone leads to $2\%$ improvement in mAP in object detection tasks.
     \item We demonstrate that CL learns different features, with coarser features in early layers and finer features in later layers whereas end-to-end learned features have more evenly distributed granulometry across layers.
\end{itemize}

\section{Methodology}
In this section, we introduce our proposed methodology. Firstly, we describe our technical contribution. Secondly, we describe different techniques in feature visualisation. Thirdly, we briefly describe the YOLO framework~\cite{Redmon2016YouOL} and how it performs bounding box prediction in object detection tasks. Lastly, we introduce our quantification metric and datasets used in our experiments. 

\subsection{Deep Cascade Learning in Feature Localisation}
One of the important technical contributions of our work is to train the deep neural network from scratch via CL, then perform feature visualisation at different layers and investigate the differences in the feature representations with respect to the labelled bounding box of interest. We perform an identical experimental setup for E2E-trained models. This is partially done by retraining classifiers tapped after every convolutional layer. Our experimental result suggests that E2E-trained models are not localised to the bounding box of interest.
Section~\ref{sec:feature_visualisation} details the feature visualisation methods we adopt in our experiments. Despite the methodology being straightforward, we make an important observation that CL produces high-quality visual explanations compared to identical architectures trained via E2E learning. Our localisation experiment quantitatively demonstrates that feature saliency generated by CL highly overlaps with the region of interest annotated by domain experts.
Furthermore, we propose to use CL in DCNN training as an effective bounding box regressor. Our experimental result suggests that DCNN backbone trained via CL improves performance in object detection. Section~\ref{sec:bboxyolo} includes the methodology details of the bounding box prediction method.

\subsection{Feature Visualisation}
\label{sec:feature_visualisation}
Sometimes it is not sufficient to report and be satisfied with strong performance measures on general datasets when delivering care for patients~\cite{Esteva2021}. It requires a deep understanding of which cases the model has made a good performance and which circumstances it fails. Feature visualisation provides a visual explanation by plotting salient images showing the most contributing pixel location~\cite{simonyan2014deep,Selvaraju2020}, or selecting image patches that are potentially interpretable by a model trained via perturbed images~\cite{10.1145/2939672.2939778}.

\subsubsection{Saliency Map}
\textbf{Saliency map}~\cite{simonyan2014deep} measures sensitivity for individual pixels, given an input image $I$ on the final prediction. This is achieved by taking the gradient of the class score ($S_c$) with respect to the input image itself:
\begin{equation}
w=\frac{\partial S_{c}}{\partial I}
\end{equation}

The result will give us a contribution map of the degree to which a pixel contributed to that class score. This gives us insight into what the network is focusing on with respect to the input image for each particular class prediction. 

\subsubsection{Grad-CAM}
The \textbf{Grad-CAM}~\cite{Selvaraju2020} method generates a heat-map of the input pixels, telling us where the model is looking at to make a particular prediction. Grad-CAM considers how a change in a particular location $i$, $j$, in the activation map $A^k$, creates a change in the class activation $y_c$ by computing this gradient (Equation~\ref{eqa:importance_score}). This is accumulated by summing the values over the entire activation map indexed by $k$ to give $\alpha_k^c$. The scalar $\alpha_{k}^{c}$ represents \textit{neuron importance} for the $k^{th}$ feature map and class $c$. Finally, $L_{\mathrm{Grad}-\mathrm{CAM}}$ is computed using Equation~\ref{eqa:gradcam}, where $Z$ denotes the total number of pixels in the feature map. Equation~\ref{eqa:gradcam} accumulates the neuron importance over all the activation maps, followed by the ReLU non-linearity to remove the negative components. $\alpha_{k}^{c} < 0$ implies that a change in $A^k$ will decrease prediction score $y^c$, which should be avoided as those feature maps that improve the prediction are of interest~\cite{Selvaraju2020}, hence the ReLU:

\begin{equation}
\alpha_{k}^{c}=\frac{1}{Z} \sum_{i} \sum_{j} \frac{\partial y^{c}}{\partial A_{i j}^{k}}
\label{eqa:importance_score}
\end{equation}

\begin{equation}
L_{\mathrm{Grad}-\mathrm{CAM}}^{c}=\operatorname{ReLU} \left(\sum_{k} \alpha_{k}^{c} A^{k}\right)
\label{eqa:gradcam}
\end{equation}

\subsubsection{LIME}
\begin{figure}[!htb]
    \centering
    \includegraphics[width=0.7\textwidth]{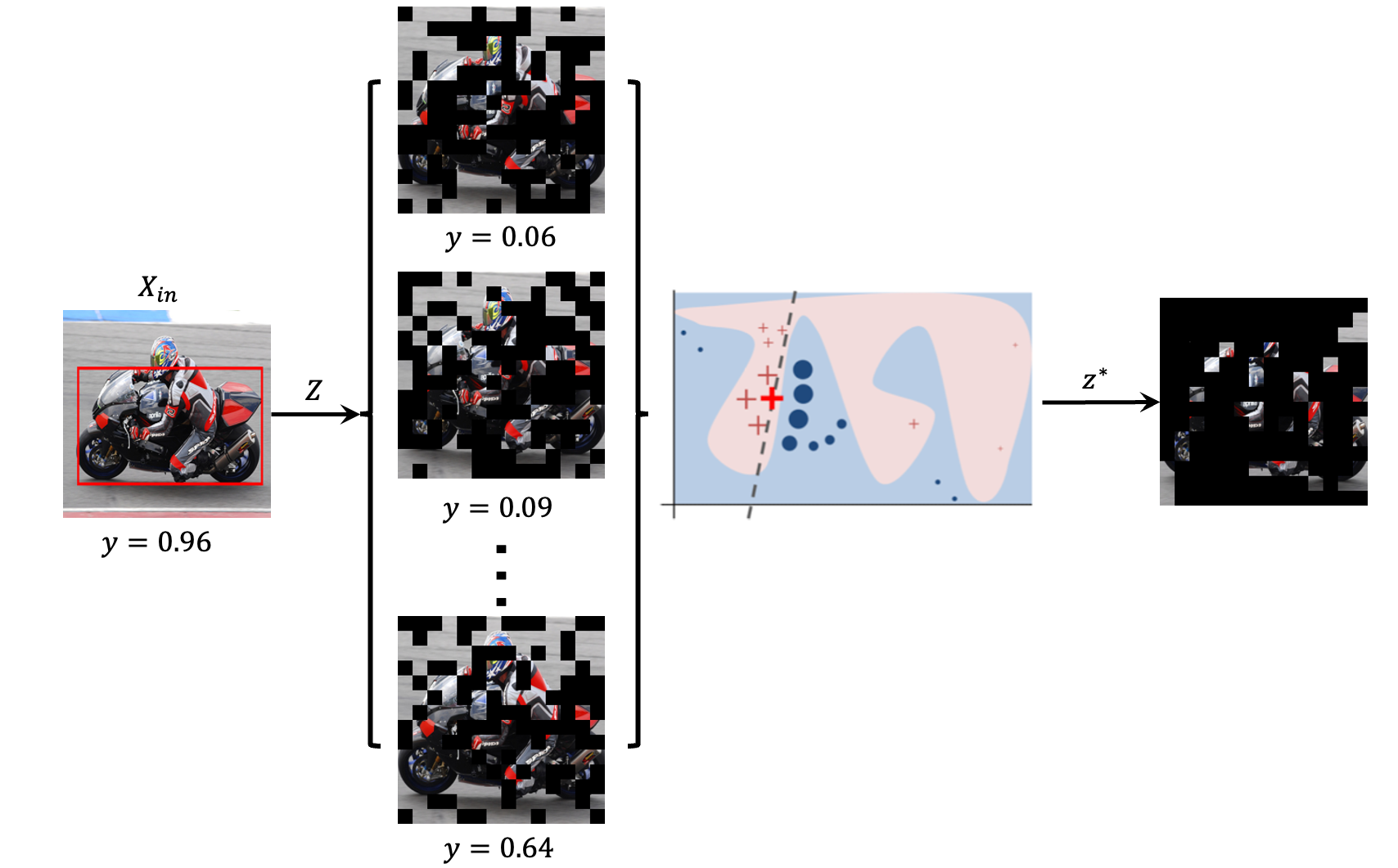}
    \caption{Illustration of the LIME framework. $X_{in}$ is input image; $y$ is confidence score output by model. They introduce a binarized ``intermediate representation'' $\boldsymbol{z}$ to represent the existence of certain image patches.}
    \label{fig:CL_explaination_pertube}
\end{figure}

Local Interpretable Model-agnostic Explanations (LIME)~\cite{10.1145/2939672.2939778} generates an occluded version of the image as a visual explanation. This is achieved by randomly perturbing the image patch (allocated by super-pixel) and training simple classifiers (e.g. ridge regression) using prediction score from the model to be explained~\cite{10.1145/2939672.2939778}. By performing the feature selection on a simple classier, super-pixels that contribute largely to final predictions are found. Figure~\ref{fig:CL_explaination_pertube} shows an illustration of the LIME framework~\cite{10.1145/2939672.2939778}.

\subsection{Bounding Box Prediction Via YOLO}
\label{sec:bboxyolo}

You Only Look Once (YOLO)~\cite{Redmon2016YouOL, Redmon2018YOLOv3AI} framework aims to predict all bounding boxes by inputting images once. It splits a given image into an $S\times S$ grid, and predicts the bounding box coordinate depending on which grid the center of the bounding box is located. It optimizes the following bound-box regression loss:
\begin{equation}
\begin{aligned}
L_{BBox} = & \sum_{i=0}^{S^2} \sum_{j=0}^B \mathbbm{1}_{i j}^{\text {obj }}\left[\left(x_i-\hat{x}_i\right)^2+\left(y_i-\hat{y}_i\right)^2\right] \\
&\quad+ \sum_{i=0}^{S^2} \sum_{j=0}^B \mathbbm{1}_{i j}^{\text {obj }}\left[\left(\sqrt{w_i}-\sqrt{\hat{w}_i}\right)^2+\left(\sqrt{h_i}-\sqrt{\hat{h}_i}\right)^2\right]
\label{eqa:bbox_reg}
\end{aligned}
\end{equation}
where $x,y,w,h$ represent the two-dimensional object's center coordinate, width and height, respectively. Final loss is calculated by iterating through all grids $S$ and object bounding boxes $B$. To predict coordinate information, it is normal to modify the output layer or add extra detection head~\cite{Redmon2016YouOL, Redmon2018YOLOv3AI}. However, the network backbone remains the same. In order to implement CL for the object detection task, we consider a two-step approach. First, pre-training the backbone network via CL on image classification tasks. Second, perform bounding-box regression via Equation~\ref{eqa:bbox_reg}. In our experiments, we consider two backbone network structures, which are a simple $6$ layer DCNN model and a $53$ layer DarkNet~\cite{Redmon2016YouOL}.

\subsection{Metrics}
To quantify the localisation ability, we use the {\it Intercept Over Union} (IOU) metric:
\begin{equation}
IOU=\frac{\operatorname{area}\left(B_{p} \cap B_{gt}\right)}{\operatorname{area}\left(B_{p} \cup B_{gt}\right)}
\end{equation}
where $B_{p}$ denotes the binarized saliency map. For the thresholding process, we use a fixed percentile instead of a constant value, ensuring a fair comparison. This results in binarized saliency maps that all have the same degree of pixel covering but are different in distribution. $B_{gt}$ denotes the binarized ground truth bounding box, where regions inside the box are \textit{True}. To quantify the model's overall localisation ability, we define {\it Localisation Accuracy} by measuring the fraction of instances that satisfy $IOU > 0.2$. Note that the LIME framework explains the decision at the patch level. However, we are merely interested in part of the patch that overlaps with the bounding box. Therefore, we measure mainly the degree of overlap by counting the number of pixels that are inside the bounding box.

For the object detection task, we evaluate our model performance using {\it mean Average Precision (mAP)}~\cite{Lin2014MicrosoftCC} and {\it mean Intersection over Union (mIOU)}. We are using both single and multiple IOU thresholds to measure mAP. For a single IOU threshold, we select $IOU = 0.5$ and $0.75$. For multiple IOU thresholds, we use the mean of $10$ IOU thresholds, from $0.5$ to $0.95$ with step size $0.05$.

\subsection{Datasets}
We show our method has improved localisation ability in both natural image and medical domains. Specifically, for the natural image domain, we choose Pascal VOC~\cite{Everingham10} which includes $11,530$ natural images in 20 classes and $27,450$ object ROI since multiple objects exist. For chest X-ray images we use the ChestX-ray8~\cite{Wang2017} dataset, where $987$ chest X-ray images are provided with board certified medic annotations of the correct location of the anomaly.

\section{Results}
Next, we present experimental results on the two varied datasets of natural and medical images. 

\subsection{Feature Visualisation}
In Figure~\ref{fig:gradcam_chestxray}, we visualise the dominant features learned by the network across various layers using the Grad-CAM~\cite{Selvaraju2020} saliency map. We observe a large gap between CL (top row) and E2E (bottom row) features on the chest X-ray data. CL features are often more visually localised with respect to the bounding box. Similar phenomena are observed by only visualising the gradient signals across various layers as illustrated in Figure~\ref{fig:CL_saliency}. These results suggest that the gradient signal plays an important role in generating a qualitative visualisation. Next, we quantify this effect by considering the IOU and localisation accuracy over both datasets.

\subsection{Feature Localisation via Grad-CAM and Saliency Map}

\begin{figure}[!htb]
    \centering
    \begin{tabular}{c}
    \includegraphics[width=0.2\textwidth]{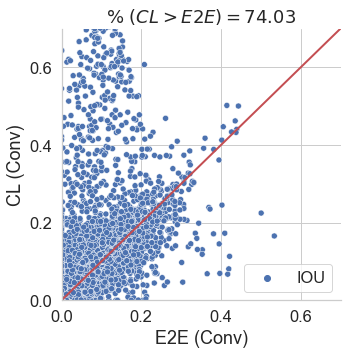}\\
    \small (a) IOU (Pascal)
    \end{tabular}
    \begin{tabular}{c}
    \includegraphics[width=0.2\textwidth]{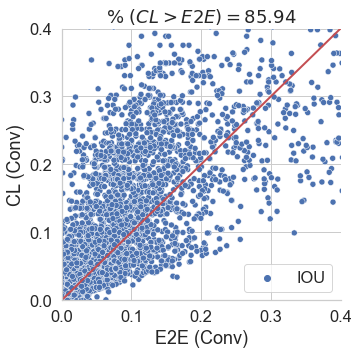}\\
    \small (b) IOU (Chest X-ray)
    \end{tabular}
    \begin{tabular}{c}
    \includegraphics[width=3.7cm,height=3cm]{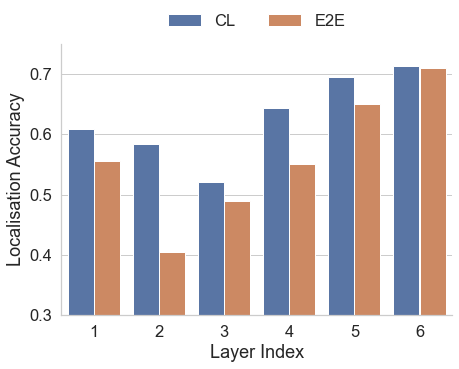}\\
    \small (c) L-ACC (Pascal)
    \end{tabular}
    \begin{tabular}{c}
    \includegraphics[width=3.7cm,height=3cm]{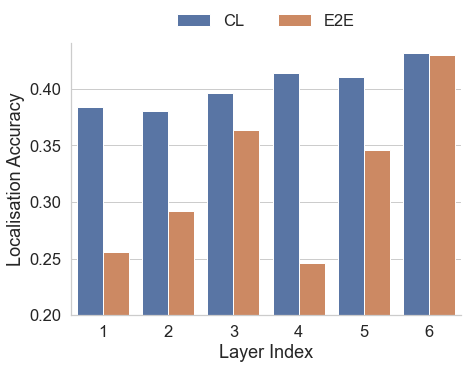}\\
    \small (d) L-ACC (Chest X-ray)
    \end{tabular}

    \caption{a-b): Scattering plot of IOU between the manual annotation and saliency maps. The experiment was conducted on both natural images (Pascal) and medical datasets (Chest X-ray). c-d): IOU between the \textit{Grad-CAM} and bounding boxes, over varied learning method layers. }
    \label{fig:localisation_acc}
\end{figure}

Figure~\ref{fig:localisation_acc} shows the scatter plots of IOU computed over (a) $2000$ Pascal images and (b) $987$ chest X-ray images. Each data point corresponds to an image, with the IOU of the network trained with E2E presented on the x-axis, and CL on the y-axis. The majority of the images have more localised features (higher IOU) with CL as opposed to E2E, with $74\%$ on the Pascal dataset and $86\%$ with the Chest x-ray dataset. The localisation accuracy is further plotted over the layers of the network in Figure~\ref{fig:localisation_acc}(c) and (d) demonstrating the superiority in feature localisation for networks trained via CL.

\subsection{Feature Localisation via LIME Framework}
\begin{figure}[!htb]
    \centering
    \includegraphics[width=0.5\textwidth]{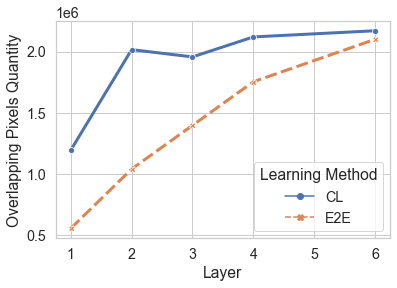}
    \caption{Localisation performance comparing CL and E2E learning method using LIME framework.}
    \label{fig:lime_test}
\end{figure}

In this section, we evaluate CL localisation performance using LIME~\cite{10.1145/2939672.2939778}. LIME requires learning multiple simple learners which creates extra complexity. But it does not require gradient information, which differs from a gradient-based method such as Grad-CAM~\cite{Selvaraju2020} and saliency map~\cite{simonyan2014deep}. In Figure~\ref{fig:lime_test}, we show that CL produces meaningful features by measuring the degree of overlap between LIME output images (occluded area are treated as $0$) and the bounding box. Figure~\ref{fig:lime_test} shows localisation performance comparing CL and E2E learning methods using the LIME test. CL learned features consistently outperform the E2E features in all layers, with the largest improvement found at the second layer.

\subsection{CL Improves Region Proposal}
\begin{figure}[!htb]
    \centering
    \includegraphics[width=0.5\textwidth]{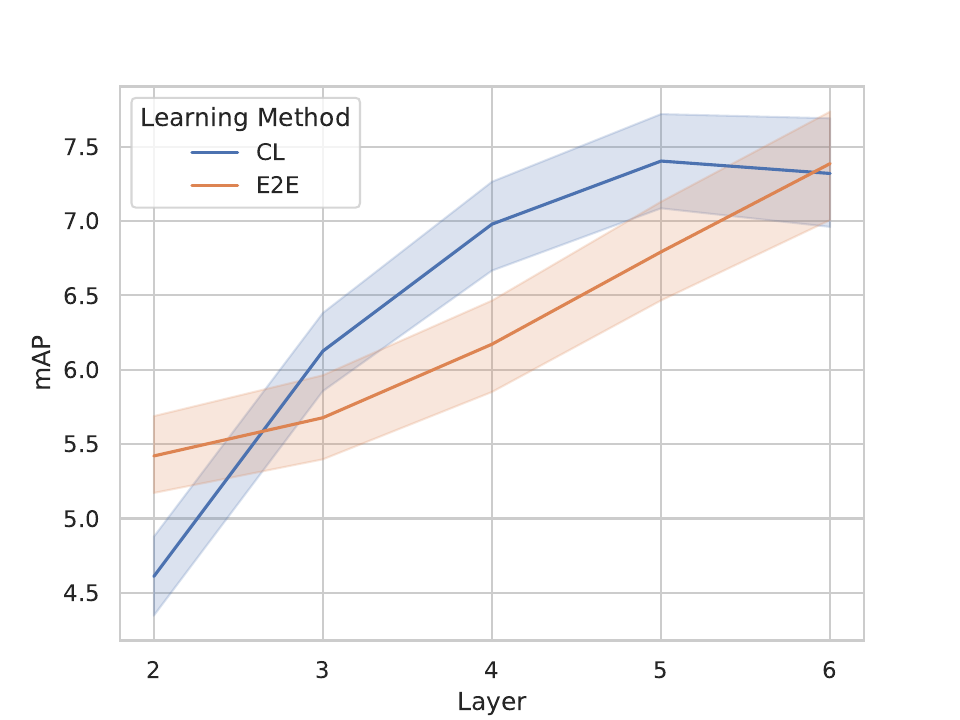}
    \caption{CL achieves better performance with the largest difference in mAP at layer $4$. For each layer, we re-train CL in three different random seeds. The shaded area denotes the standard deviation.}
    \label{fig:map_varying_methods}
\end{figure}

We show that training a backbone network via CL improves region proposal in bounding box prediction. We adopt a network trained via CL as an effective bounding box regressor. We keep feature layers frozen and retrain one added convolutional layer to optimize the bounding box regression loss. The loss was first introduced in YOLOv1 framework~\cite{Redmon2016YouOL}. The layerwise comparison results are shown in Figure~\ref{fig:map_varying_methods}. We found CL achieves the best overall quality of region proposal with the largest difference at layer $4$ compared to an identical network trained via E2E. Notice that we are not directly competing with the state-of-the-art model, but we claim using the CL training scheme improves overall performance in object detection tasks against the widely adopted E2E scheme.

We further investigate whether using a deep network backbone trained via CL could improve region proposal. Table~\ref{tab:CL_trianing_backbone} shows YOLOv3 performance on the Pascal dataset. We adopt CL to train the DarkNet-53~\cite{Redmon2018YOLOv3AI} network backbone from scratch and compare against the E2E baseline training method. To implement the CL algorithm on the DarkNet-53 architecture, we split the whole structure into multiple sub-modules. Each sub-module consists of at least one complete residual connection block. When the network architecture is fixed, the size of the sub-module is determined by the total number of splits. In the YOLOv3 framework, the output is taken from three locations among intermediate features and passed to the feature pyramid network (FPN) to improve detection for different object sizes. In our experiment, we split the network into three sub-modules and denote it as $CL_3$. This result, along with other splitting strategies to train with CL, are reported in Table~\ref{tab:CL_trianing_backbone}. We found using pre-trained features from the middle layer of the network yields the largest difference between CL and E2E. The best performance using CL feature up to middle layer improves 2\% in $mAP_{.5}$ metric compared to reusing E2E feature at same layer. When increasing the number of splits, the performance starts to decrease. This is possibly caused by overfitting since the network's learning ability is limited due to sub-module size shrinks by having larger quantities of splits.
\begin{table}[!htb]
\centering

\begin{tabular}{lllll}
\toprule
{} &             $mAP_{.5:.95:.05}$ &          $mAP_{.5}$ &         $mAP_{.75}$ \\
\midrule
$CL_{3}$  &  34.64$\pm$0.31 &  67.29$\pm$0.28 &   31.77$\pm$0.4 \\
$CL_{7}$  &   33.87$\pm$0.1 &  66.56$\pm$0.15 &   30.27$\pm$0.3 \\
$CL_{23}$ &  33.34$\pm$0.28 &  65.68$\pm$0.23 &  29.63$\pm$0.62 \\
$E2E$     &  33.41$\pm$0.18 &  65.3$\pm$0.24 &  30.07$\pm$0.43 \\
\bottomrule
\end{tabular}

\caption{Comparing CL and E2E training method to train network backbone in YOLOv3 framework~\cite{Redmon2018YOLOv3AI}. All CL and E2E are using DarkNet-53~\cite{Redmon2018YOLOv3AI} architecture. The lower subscript denotes the total number of splits in CL training. }
\label{tab:CL_trianing_backbone}
\end{table}
\subsection{Quantifying Coarse-to-Fine Features Representation}
\begin{figure}[!htb]
    \centering
    \begin{tabular}{c}
    \includegraphics[width=5cm,height=4cm]{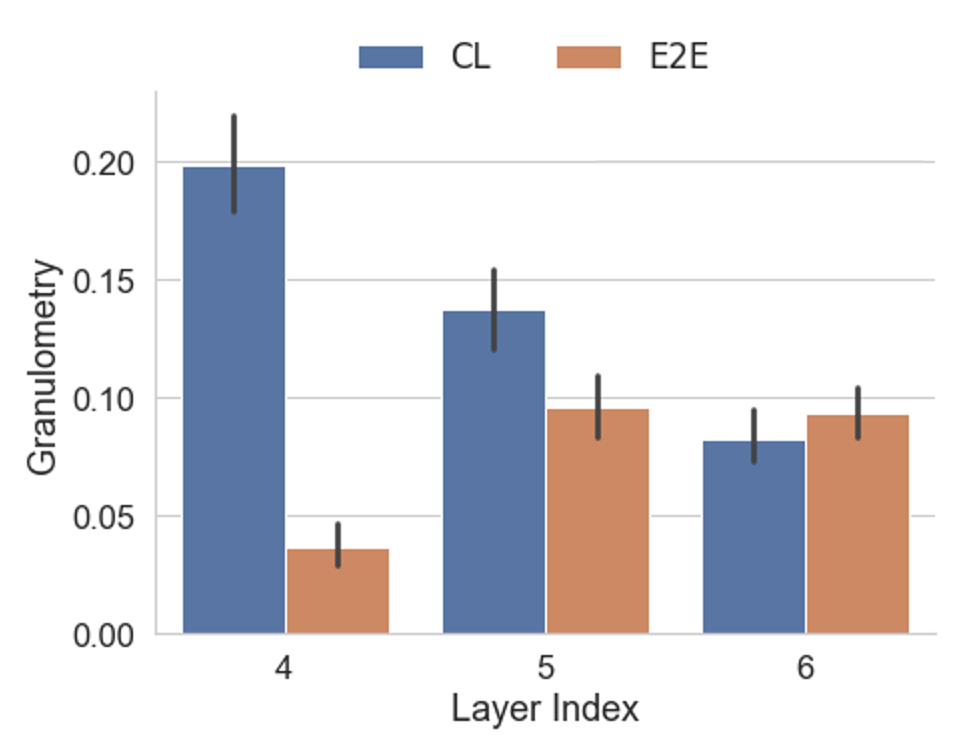}\\
    \small (a) Pascal
    \end{tabular}
    \begin{tabular}{c}
    \includegraphics[width=5cm,height=4cm]{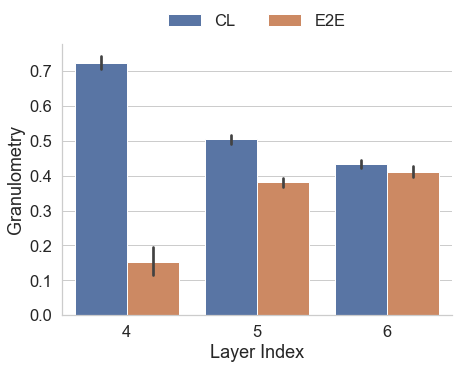}\\
    \small (b) Chest X-ray
    \end{tabular}
    \caption{Granulometry measure comparing CL and E2E learning methods on different layers (layer $1-3$ as the inconsistency observed in early layers). a) Pascal; b) chest X-ray}
    \label{fig:granulometry}
\end{figure}

Granulometry analysis~\cite{dougherty1989image} on the generated saliency maps quantitatively demonstrates the coarse-to-fine feature representation. The higher granulometry represents the feature activation (indicated as the irregular red patch in Figure~\ref{fig:gradcam_chestxray}) are coarser, finer detail is learned if granulometry has a low value. Figure~\ref{fig:granulometry} quantitatively analyze using granulometry to measure CL and E2E feature representation. We conclude that CL is learning coarser feature representation at early layers and finer at later layers. On the contrary, E2E has more evenly distributed granulometry across the layers. These results strengthen the argument for CL learning optimal feature representation as we demonstrate that early layers in the network are learning coarser features while later layers are learning more fine-grained features. 

\subsection{Visualisation for Small Object Bounding Box}
\begin{figure}[!htb]
  \centering
  \includegraphics[width=0.6\textwidth]{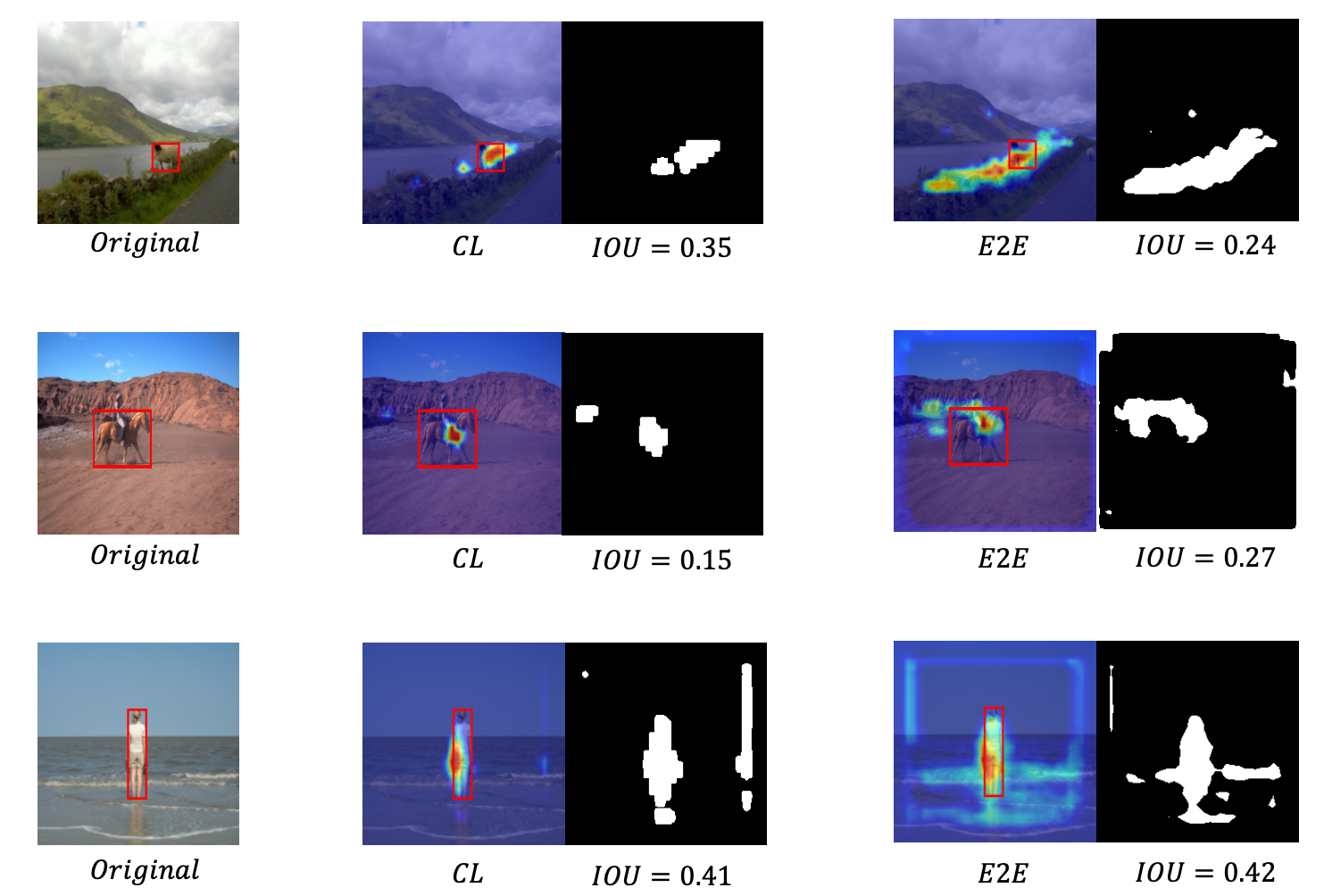}
  \caption{Visualisation of data instance via Grad-CAM generated from two methods. left: Original image and its bounding box; middle: CL ; right: E2E. The IOU value associated with each binarized saliency map is shown on the right side.}
  \label{fig:failcase}
\end{figure}

In Figure~\ref{fig:failcase}, we visualise instances with a relatively small bounding box. By visualising the instance and its corresponding binary mask, we observed that CL is able to generate a localised heatmap for the small object of interest with a high-quality salient image. On the other hand, E2E tend to generate salient images that are activated in a large region, result in an ambiguous localisation.

\section{Conclusion}
In this work we investigate the localisation of features across learning paradigms. Our systematic evaluation across various feature visualisation methods and datasets show that E2E training, which has been widely considered by the machine learning community, is limited to localising discriminative features across multiple network layers.
We found network trained via CL is more localised to the region of interest annotated by domain experts. We show that CL's superior localisation ability leads to an improvement in object detection tasks. 

\bibliographystyle{unsrtnat}
\bibliography{references}  

\section{Appendix}

\subsection{More Results}

\begin{figure}[!htb]
    \centering
    \begin{tabular}{c}
    \includegraphics[width=0.2\textwidth]{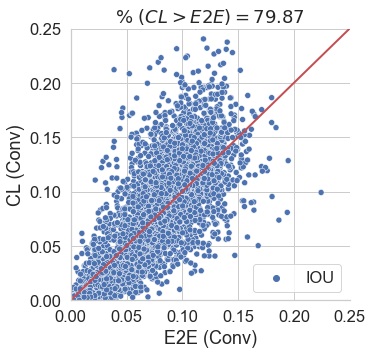}\\
    \small (a) IOU (Pascal)
    \end{tabular}
    \begin{tabular}{c}
    \includegraphics[width=0.2\textwidth]{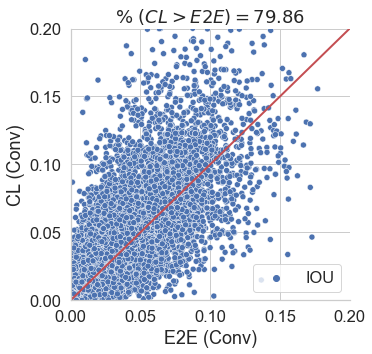}\\
    \small (b) IOU (Chest X-ray)
    \end{tabular}
    \begin{tabular}{c}
    \includegraphics[width=3.7cm,height=3cm]{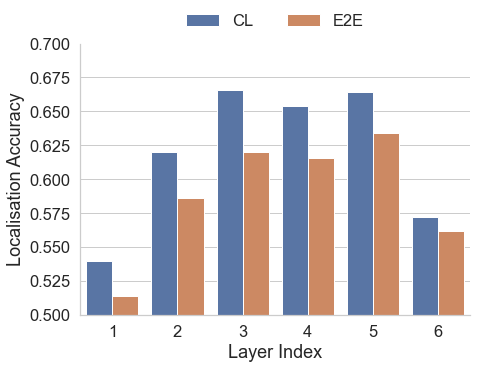}\\
    \small (c) L-ACC (Pascal)
    \end{tabular}
    \begin{tabular}{c}
    \includegraphics[width=3.7cm,height=3cm]{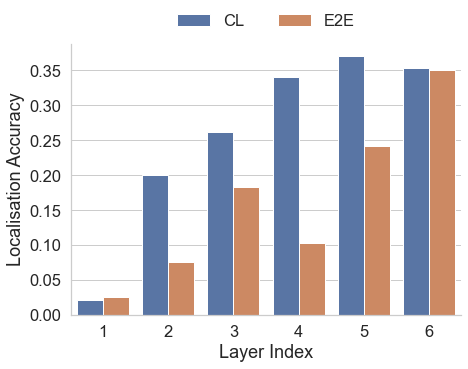}\\
    \small (d) L-ACC (Chest X-ray)
    \end{tabular}
    
    \caption{a-b): Scattering plot of IOU between the manual annotation and saliency maps. The experiment was conducted on both natural image (pascal) and medical dataset (chest X-ray) c-d): IOU between the \textit{saliency maps} output and bounding boxes, over different layers.}
    \label{fig:localisation_acc_saliencymap}
\end{figure}

Figure~\ref{fig:localisation_acc_saliencymap} shows scatter plots of IOU and localisation accuracy. Quantifying the localisation ability of CL via \textit{Saliency Map}~\cite{simonyan2014deep}. Align with the result in Figure~\ref{fig:localisation_acc}, CL learning scheme consistently improves localisation over E2E learning scheme.

\begin{figure}[!htb]
    \begin{tabular}{c}
    \includegraphics[width=8cm,height=4cm]{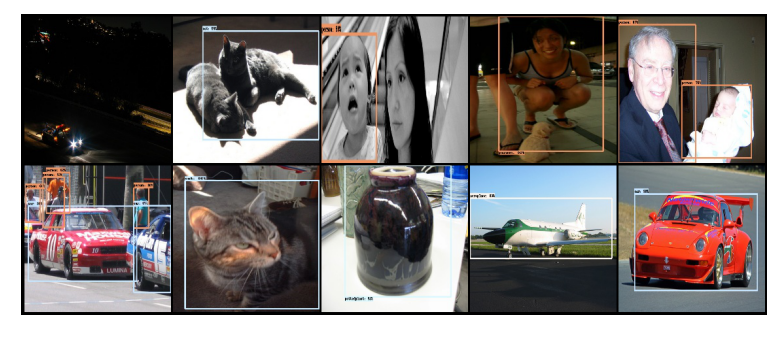}\\
    \small (a) CL
    \end{tabular}
    \begin{tabular}{c}
    \includegraphics[width=8cm,height=4cm]{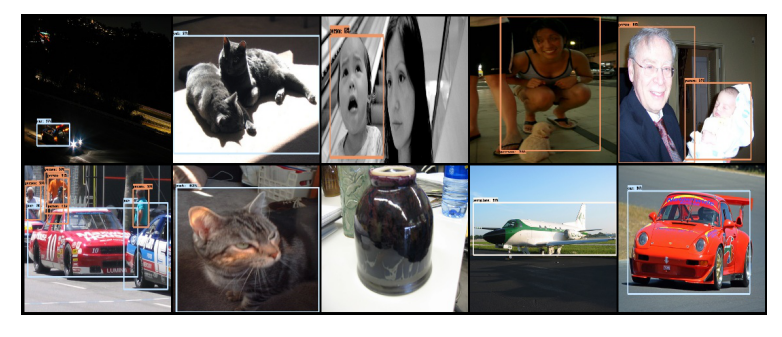}\\
    \small (b) E2E
    \end{tabular}

    \caption{Qualitative Results. YOLO prediction on random test sample from Pascal dataset. Comparing CL and E2E training scheme to train network backbone.}
    \label{fig:CL_visual_bbox}
\end{figure}

Figure~\ref{fig:CL_visual_bbox} provides qualitative analysis by visualising bounding-box prediction for some randomly selected images. We notice CL is able to predict a precise bounding box location. On the other hand, E2E fails to generate the bounding box (e.g. second row, image of jar) or generate imprecise location (e.g. first row, image of two cats).

\end{document}